\documentclass[letterpaper]{article}
\usepackage{aaai}
\usepackage{times}
\usepackage{helvet}
\usepackage{courier}
\usepackage{url}
\usepackage{amsmath}
\usepackage{amsbsy}
\usepackage{amssymb}
\usepackage{amsopn}
\usepackage{stmaryrd}
\usepackage{graphicx}
\usepackage{multirow}
\usepackage{rotating}
\begin{document}
%
\title{A Bayesian Model for Plan Recognition in RTS Games applied to StarCraft}
\author{Gabriel Synnaeve and Pierre Bessi\`{e}re\\
LPPA, Coll\`{e}ge de France, UMR7152 CNRS\\
11 place Marcelin Berthelot, 75231 Paris Cedex 05, France\\
\texttt{gabriel.synnaeve@gmail.com}\ \ \texttt{pierre.bessiere@imag.fr}
}
\maketitle
\begin{abstract}
\begin{quote}
The task of keyhole (unobtrusive) plan recognition is central to adaptive game AI. ``Tech trees'' or ``build trees'' are the core of real-time strategy (RTS) game strategic (long term) planning. This paper presents a generic and simple Bayesian model for RTS build tree prediction from \textit{noisy} observations, which parameters are learned from \textit{replays} (game logs). This \textit{unsupervised} machine learning approach involves minimal work for the game developers as it leverage players' data (common in RTS). We applied it to StarCraft\footnote{StarCraft and its expansion StarCraft: Brood War are trademarks of Blizzard Entertainment$^{\mathrm{TM}}$} and showed that it yields high quality and robust predictions, that can feed an adaptive AI.
\end{quote}
\end{abstract}

\section{Introduction}
In a RTS, players need to gather resources to build structures and military units and defeat their opponents. To that end, they often have \textit{worker units} than can gather resources needed to build \textit{workers}, \textit{buildings}, \textit{military units} and \textit{research upgrades}. Resources may have different uses, for instance in StarCraft: minerals are used for everything, whereas gas is only required for advanced buildings or military units, and technology upgrades. The military units can be of different types, any combinations of ranged, casters, contact attack, zone attacks, big, small, slow, fast, invisible, flying... Units can have attacks and defenses that counter each others as in rock-paper-scissors. Buildings and research upgrades define technology trees (precisely: directed acyclic graphs). Tech trees are tied to strategic planning, because they put constraints on which units types can be produced, when and in which numbers, which spells are available and how the player spends her resources.

Most real-time strategy (RTS) games AI are either not challenging or not fun to play against. They are not challenging because they do not adapt well dynamically to different strategies (long term goals and army composition) and tactics (army moves) that a human can perform. They are not fun to play against because they cheat economically, gathering resources faster, and/or in the intelligence war, bypassing the fog of war. We believe that creating AI that adapt to the strategies of the human player would make RTS games AI much more interesting to play against.

We worked on StarCraft: Brood War, which is a canonical RTS game, as Chess is to board games. It had been around since 1998, it has sold 10 millions licenses and was the best competitive RTS for more than a decade. There are 3 factions (Protoss, Terran and Zerg) that are totally different in terms of units, tech trees and thus gameplay styles. StarCraft and most RTS games provide a tool to record game logs into \textit{replays} that can be re-simulated by the game engine and watched to improve strategies and tactics. All high level players use this feature heavily either to improve their play or study opponents style. Observing replays allows players to see what happened under the fog of war, so that they can understand timing of technologies and attacks and find clues/evidences leading to infer the strategy as well as weak points (either strategic or tactical). We used this replay feature to extract players actions and learn the probabilities of tech trees to happen at a given time.

\begin{figure}[htp]
\centerline{\includegraphics[width=0.95\columnwidth]{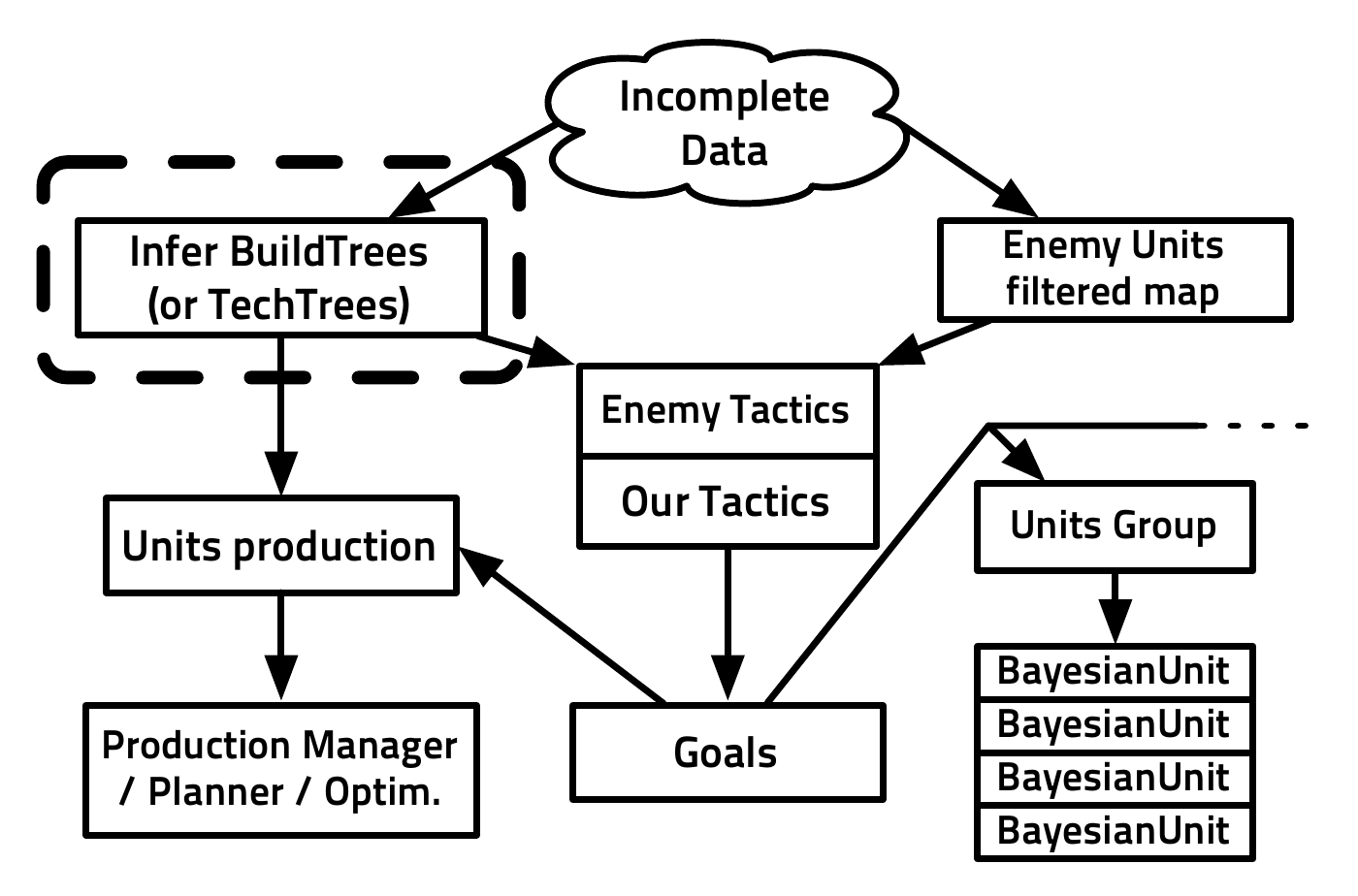}}
\caption{Data flow of the free software StarCraft robotic player \textsc{BroodwarBotQ}. In this paper, we only deal with the upper left part (in a dotted line).}
\label{bbq_dataflow}
\end{figure}

In our model, we used the buildings part of tech trees because buildings can be more easily viewed than units when fog of war is enforced, and our main focus was our StarCraft bot implementation (see Figure~\ref{bbq_dataflow}), but nothing hinders us to use units and upgrades as well in a setting without fog of war (commentary assistant or game AI that cheat).

\section{Background}
\subsection{Related Works}
This work was encouraged by the reading of \citeauthor{weber} \shortcite{weber} Data Mining Approach to Strategy Prediction and the fact that they provided their dataset. They tried and evaluated several machine learning algorithms on replays that were labeled with strategies (supervised learning).

There are related works in the domains of opponent modeling \cite{HsiehS08,SchaddBS07,OBRecog}. The main methods used to these ends are case-based reasoning (CBR) and planning or plan recognition \cite{LTW,CBR_Planning,OntanonCBR,HTNPlanning,Ramirez}. There are precedent works of Bayesian plan recognition \cite{BMPR}, even in games with \cite{BayesianRecog} using dynamic Bayesian networks to recognize a user's plan in a multi-player dungeon adventure. Also, \citeauthor{Chung05} \shortcite{Chung05} describe a Monte-Carlo plan selection algorithm applied to Open RTS.

\citeauthor{LTW} \shortcite{LTW} used CBR to perform dynamic plan retrieval extracted from domain knowledge in Wargus (Warcraft II clone). \citeauthor{CBR_Planning} \shortcite{CBR_Planning} base their real-time case-based planning (CBP) system on a plan dependency graph which is learned from human demonstration. In \cite{OntanonCBR,PlanRetrieval}, they use CBR and expert demonstrations on Wargus. 
They improve the speed of CPB by using a decision tree to select relevant features. \citeauthor{HsiehS08} \shortcite{HsiehS08} based their work on CBR and \citeauthor{LTW} \shortcite{LTW} and used StarCraft replays to construct states and building sequences. Strategies are choices of building construction order in their model. 

\citeauthor{SchaddBS07} \shortcite{SchaddBS07} describe opponent modeling through hierarchically structured models of the opponent behaviour and they applied their work to the Spring RTS (Total Annihilation clone). \citeauthor{HTNPlanning} \shortcite{HTNPlanning} use hierarchical task networks (HTN) to model strategies in a first person shooter with the goal to use HTN planners. \citeauthor{OBRecog} \shortcite{OBRecog} improve the probabilistic hostile agent task tracker (PHATT \cite{PHATT}, a simulated HMM for plan recognition) by encoding strategies as HTN.

The work described in this paper can be classified as probabilistic plan recognition. Strictly speaking, we present model-based machine learning used for prediction of plans, while our model is not limited to prediction. The plans are build trees directly learned from the replays (unsupervised learning).

\subsection{Bayesian Programming}
Probability is used as an alternative to classical logic and we transform incompleteness (in the experiences, the perceptions or the model) into uncertainty \cite{Jaynes}. We introduce Bayesian programs (BP), a formalism that can be used to describe entirely any kind of Bayesian model, subsuming Bayesian networks and Bayesian maps, equivalent to probabilistic factor graphs \cite{Diard03}. There are mainly two parts in a BP, the \textbf{description} of how to compute the joint distribution, and the \textbf{question(s)} that it will be asked. 

The description consists in explaining the relevant \textit{variables} $\{X^1,\dots,X^n\}$ and explain their dependencies by \textit{decomposing} the joint distribution $P(X^1\dots X^n | \delta, \pi)$ with existing preliminary knowledge $\pi$ and data $\delta$. The \textit{forms} of each term of the product specify how to compute their distributions: either parametric forms (laws or probability tables, with free parameters that can be learned from data $\delta$) or recursive questions to other Bayesian programs.

Answering a question is computing the distribution $P(Searched | Known)$, with $Searched$ and $Known$ two disjoint subsets of the variables. 
$P(Searched | Known) $
$$ = \frac{\sum_{Free}P(Searched,\ Free,\ Known)}{P(Known)}$$ $$ = \frac{1}{Z}\times \sum_{Free} P(Searched,\ Free,\ Known)$$

General Bayesian inference is practically intractable, but conditional independence hypotheses and constraints (stated in the description) often simplify the model. Also, there are different well-known approximation techniques, for instance Monte Carlo methods 
and variational Bayes \cite{Beal}. In this paper, we will use only simple enough models that allow complete inference to be computed in real-time.

\begin{small}
\begin{eqnarray*}
BP
\begin{cases}
Desc.
    \begin{cases}
    Spec. (\pi)
        \begin{cases}
        Variables\\
        Decomposition\\
        Forms\ (Parametric\ or\ Program)
        \end{cases}\\
    Identification\ (based\ on\ \delta)
    \end{cases}\\
Question
\end{cases}
\end{eqnarray*}
\end{small}
For the use of Bayesian programming in sensory-motor systems, see \cite{PRDMSMS}. For its use in cognitive modeling, see \cite{Colas10}. For its first use in video games (first person shooter gameplay, Unreal Tournament), see \cite{LeHy04}.

\section{Methodology}
\subsection{Build/Tech Tree Prediction Model}
The outline of the model is that it infers the distribution on (probabilities for each of) our opponent's build tree from observations, which tend to be very partial due to the fog of war. From what is common (conditionally of observations) and the hierarchical structure of a build tree, it diminishes or raises their probabilities.
Our predictive model is a Bayesian program, it can be seen as the ``Bayesian network'' represented in Figure~\ref{BNPrediction}. It is a generative model and this is of great help to deal with the parts of the observations' space where we do not have too much data (RTS games tend to diverge from one another as the number of possible actions grow exponentially). Indeed, we can model our uncertainty by putting a large standard deviation on too rare observations and generative models tend to converge with fewer observations than discriminative ones \cite{Jordan}. Here is the description of our Bayesian program:

\begin{figure}[htp]
\centerline{\includegraphics[width=0.6\columnwidth]{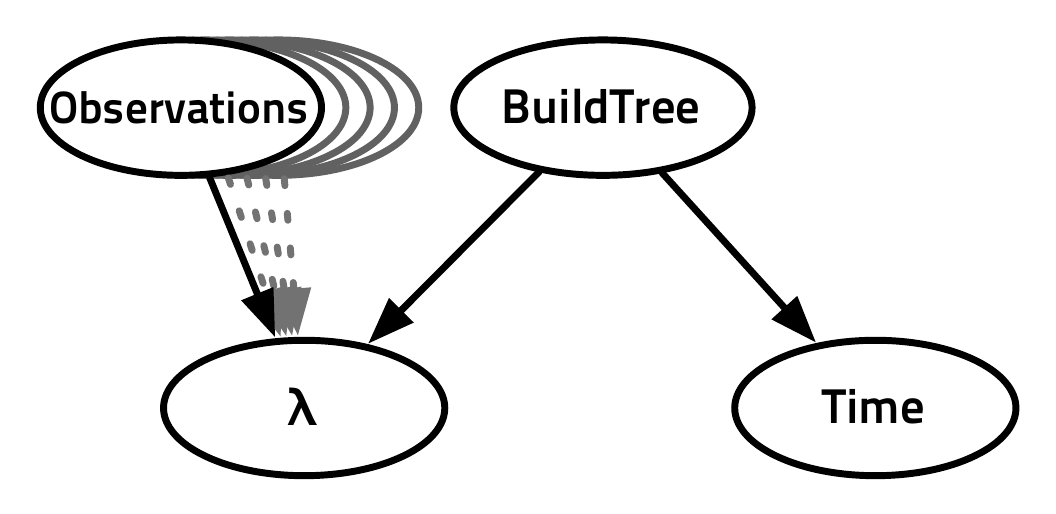}}
\caption{Graphical representation of the build tree prediction Bayesian model}
\label{BNPrediction}
\end{figure}

\subsubsection{Variables}
\begin{itemize}
\item $BuildTree \in \{\emptyset, \{building_1\}, \{building_2\}, \{building_1\wedge building_2\}, \dots\}$: all the possible building trees for the given race. For instance $\{pylon, gate\}$ and $\{pylon, gate, core\}$ are two different $BuildTrees$.
\item Observations: $O_{i \in \llbracket 1\dots N \rrbracket} \in \{0, 1\}$, $O_k$ is $1/true$ if we have seen (observed) the $k$th building (it can have been destroyed, it will stay ``seen'').
\item $\lambda \in \{0, 1\}$: coherence variable (restraining $BuildTree$ to possible values with regard to $O_{1:N}$)
\item $Time$: $T \in \llbracket 1\dots P \rrbracket$, time in the game (1 second resolution).
\end{itemize}

At first, we generated all the possible (according to the game rules) $BuildTree$ values (in StarCraft, between $\approx 500$ and $1600$ depending on the race without the same building twice). We observed that a lot of possible $BuildTree$ values are too absurd to be performed in a competitive match and were never seen during the learning. So, we restricted $BuildTree$ to have its value in all the build trees encountered in our replays dataset and we added multiple instances of the basic unit producing buildings (gateway, barracks), expansions and supply buildings (depot, pylon, ``overlord'' as a building).
This way, there are 810 build trees for Terran, 346 for Protoss and 261 for Zerg (learned from $\approx 3000$ replays for each race).

\subsubsection{Decomposition}
The joint distribution of our model is the following:
\begin{eqnarray*}
& & P(T, BuildTree, O_1 \dots O_N, \lambda) = \\ 
& & P(T | BuildTree).P(BuildTree)\\
& & P(\lambda | BuildTree, O_{1:N}).P(O_{1:N}) 
\end{eqnarray*}
This can also be see as Figure~\ref{BNPrediction}.

\subsubsection{Forms}
\begin{itemize}
\item $P(BuildTree)$ is the prior distribution on the build trees. It can either be learned from the labeled replays (histograms) or set to the uniform distribution, as we did.
\item $P(O_{1:N})$ is unspecified, we put the uniform distribution (we could use a prior over the most frequent observations).
\item $P(\lambda | BuildTree, O_{1:N})$ is a functional Dirac that restricts $BuildTree$ values to the ones than can co-exist with the observations.
\begin{eqnarray*}
& & P(\lambda = 1 | buildTree, o_{1:N}) \\
& = & 1\ \mathrm{if\ } buildTree \ \mathrm{can\ exist\ with\ } o_{1:N} \\
& = & 0\ \mathrm{else}
\end{eqnarray*}
A $BuildTree$ value ($buildTree$) is compatible with the observations if it covers them fully. For instance, $BuildTree=\{pylon, gate, core\}$ is compatible with $o_{\#core} = 1$ but it is not compatible with $o_{\#forge} = 1$. In other words, $buildTree$ is incompatible with $o_{1:N}$ \textit{iff} $\{o_{1:N} \backslash \{o_{1:N} \wedge buildTree\}\} \neq \emptyset$.
\item $P(T | BuildTree)$ are ``bell shape'' distributions (discretized normal distributions). There is one bell shape over $Time$ per $buildTree$. The parameters of these discrete Gaussian distributions are learned from the replays.
\end{itemize}

\subsubsection{Identification (learning)}
The learning of the $P(T | BuildTree)$ bell shapes parameters takes into account the uncertainty of the $buildTrees$ for which we have few observations. Indeed, the normal distribution $P(T|buildTree)$ begins with a high $\sigma^2$, and \textbf{not} a ``Dirac'' with $\mu$ on the seen $T$ value and $sigma=0$. This accounts for the fact that the first(s) observation(s) may be outlier(s). This learning process is independent on the order of the stream of examples, seeing point A and then B or B and then A in the learning phase produces the same result. 

\subsubsection{Questions}
The question that we will ask in all the benchmarks is:
\begin{eqnarray*}
 & P(BuildTree|T=t, O_{1:N}=o_{1:N}, \lambda = 1) \\
 & \propto P(t|BuildTree).P(BuildTree)\\
 & P(\lambda | BuildTree, o_{1:N}).P(o_{1:N})\\
\end{eqnarray*}
Note that if we see $P(BuildTree, Time)$ as a plan, asking $P(BuildTree | Time)$ for ourselves boils down to use our ``plan recognition'' mode as a planning algorithm, which could provide good approximations of the optimal goal set \cite{Ramirez}, or build orders. 

\begin{figure}[htp]
\centerline{\includegraphics[width=1.0\columnwidth]{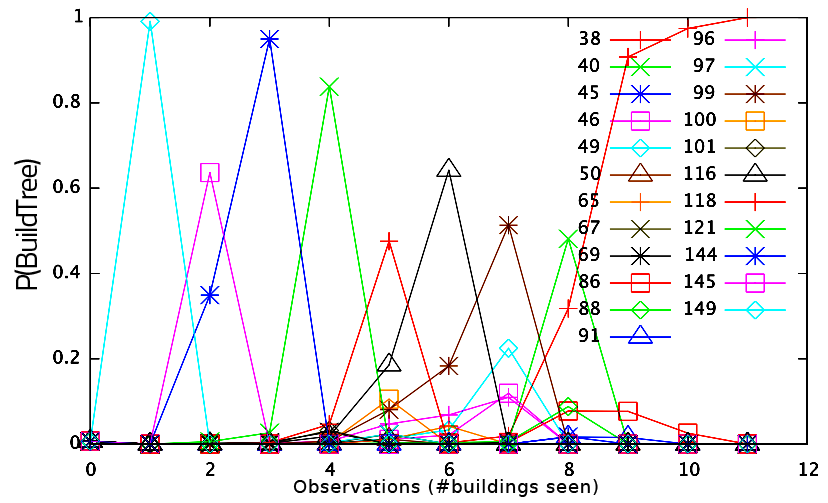}}
\caption{Evolution of $P(BuildTree|Observations...)$ in Time (seen/observed buildings on the x-axis). Only $BuildTrees$ with a probability $> 0.01$ are shown.}
\label{ttinf}
\end{figure}

\section{Results}
All the results presented in this section represents the nine match-ups (races combinations) in 1 versus 1 (duel) of StarCraft. We worked with a data-set of 8806 replays ($\approx$ 1000 per match-up) of highly skilled human players and we performed cross-validation with 9/10th of the dataset used for learning and the remaining 1/10th of the dataset used for evaluation. Performance wise, the learning part (with $\approx$ 1000 replays) takes around $0.1$ second on a 2.8 Ghz Core 2 Duo CPU (and it is serializable). Each inference (question) step takes around $0.01$ second. The memory footprint is around 3 Mb on a 64 bits machine.

\begin{table*}[ht]
\caption{Summarization of the main results/metrics, one full results set for 10\% noise}
\begin{center}
\begin{small}
\begin{tabular}{|c|c|cc|cc|cc|cc|}
\hline
  & measure & \multicolumn{2}{|c|}{$d$ for $k=0$} & \multicolumn{2}{|c|}{$k$ for $d=1$} & \multicolumn{2}{|c|}{$k$ for $d=2$} & \multicolumn{2}{|c|}{$k$ for $d=3$} \\
 noise & & \begin{scriptsize}$d(best,real)$\end{scriptsize} 
& \begin{scriptsize}$d(bt,real)*P(bt)$\end{scriptsize}
& \begin{scriptsize}best\end{scriptsize}
& \begin{scriptsize}``mean''\end{scriptsize}
& \begin{scriptsize}best\end{scriptsize}
& \begin{scriptsize}``mean''\end{scriptsize}
& \begin{scriptsize}best\end{scriptsize}
&  \begin{scriptsize}``mean''\end{scriptsize} \\
\hline
\multirow{3}{3mm}{\begin{sideways}\parbox{3mm}{0\%\ \ }\end{sideways}}
& average & 0.535 & 0.870 & 1.193 & 3.991 & 2.760 & 5.249 & 3.642 & 6.122\\
& min & 0.313 & 0.574 & 0.861 & 2.8 & 2.239 & 3.97 & 3.13 & 4.88\\
& max & 1.051 & 1.296 & 2.176 & 5.334 & 3.681 & 6.683 & 4.496 & 7.334\\
\hline
\multirow{3}{3mm}{\begin{sideways}\parbox{3mm}{10\%\ \ }\end{sideways}}
 & PvP & 0.397 & 0.646 & 1.061 & 2.795 & 2.204 & 3.877 & 2.897 & 4.693\\
 & PvT & 0.341 & 0.654 & 0.991 & 2.911 & 2.017 & 4.053 & 2.929 & 5.079\\
 & PvZ & 0.516 & 0.910 & 0.882 & 3.361 & 2.276 & 4.489 & 3.053 & 5.308\\
 & TvP & 0.608 & 0.978 & 0.797 & 4.202 & 2.212 & 5.171 & 3.060 & 5.959\\
 & TvT & 1.043 & 1.310 & 0.983 & 4.75 & 3.45 & 5.85 & 3.833 & 6.45\\
 & TvZ & 0.890 & 1.250 & 1.882 & 4.815 & 3.327 & 5.873 & 4.134 & 6.546\\
 & ZvP & 0.521 & 0.933 & 0.89 & 3.82 & 2.48 & 4.93 & 3.16 & 5.54\\
 & ZvT & 0.486 & 0.834 & 0.765 & 3.156 & 2.260 & 4.373 & 3.139 & 5.173\\
 & ZvZ & 0.399 & 0.694 & 0.9 & 2.52 & 2.12 & 3.53 & 2.71 & 4.38\\
& average & 0.578 & 0.912 & 1.017 & 3.592 & 2.483 & 4.683 & 3.213 & 5.459\\
& min & 0.341 & 0.646 & 0.765 & 2.52 & 2.017 & 3.53 & 2.71 & 4.38\\
& max & 1.043 & 1.310 & 1.882 & 4.815 & 3.45 & 5.873 & 4.134 & 6.546\\
\hline
\multirow{3}{3mm}{\begin{sideways}\parbox{3mm}{20\%\ \ }\end{sideways}}
& average & 0.610 & 0.949 & 0.900 & 3.263 & 2.256 & 4.213 & 2.866 & 4.873\\
& min & 0.381 & 0.683 & 0.686 & 2.3 & 1.858 & 3.25 & 2.44 & 3.91\\
& max & 1.062 & 1.330 & 1.697 & 4.394 & 3.133 & 5.336 & 3.697 & 5.899\\
\hline
\multirow{3}{3mm}{\begin{sideways}\parbox{3mm}{30\%\ \ }\end{sideways}}
& average & 0.670 & 1.003 & 0.747 & 2.902 & 2.055 & 3.801 & 2.534 & 4.375\\
& min & 0.431 & 0.749 & 0.555 & 2.03 & 1.7 & 3 & 2.22 & 3.58\\
& max & 1.131 & 1.392 & 1.394 & 3.933 & 2.638 & 4.722 & 3.176 & 5.268\\
\hline
\multirow{3}{3mm}{\begin{sideways}\parbox{3mm}{40\%\ \ }\end{sideways}}
& aerage & 0.740 & 1.068 & 0.611 & 2.529 & 1.883 & 3.357 & 2.20 & 3.827\\
& min & 0.488 & 0.820 & 0.44 & 1.65 & 1.535 & 2.61 & 1.94 & 3.09\\
& max & 1.257 & 1.497 & 1.201 & 3.5 & 2.516 & 4.226 & 2.773 & 4.672\\
\hline
\multirow{3}{3mm}{\begin{sideways}\parbox{3mm}{50\%\ \ }\end{sideways}}
& average & 0.816 & 1.145 & 0.493 & 2.078 & 1.696 & 2.860 & 1.972 & 3.242\\
& min & 0.534 & 0.864 & 0.363 & 1.33 & 1.444 & 2.24 & 1.653 & 2.61\\
& max & 1.354 & 1.581 & 1 & 2.890 & 2.4 & 3.613 & 2.516 & 3.941\\
\hline
\multirow{3}{3mm}{\begin{sideways}\parbox{3mm}{60\%\ \ }\end{sideways}}
& average & 0.925 & 1.232 & 0.400 & 1.738 & 1.531 & 2.449 & 1.724 & 2.732\\
& min & 0.586 & 0.918 & 0.22 & 1.08 & 1.262 & 1.98 & 1.448 & 2.22\\
& max & 1.414 & 1.707 & 0.840 & 2.483 & 2 & 3.100 & 2.083 & 3.327\\
\hline
\multirow{3}{3mm}{\begin{sideways}\parbox{3mm}{70\%\ \ }\end{sideways}}
& verage & 1.038 & 1.314 & 0.277 & 1.291 & 1.342 & 2.039 & 1.470 & 2.270\\
& min & 0.633 & 0.994 & 0.16 & 0.79 & 1.101 & 1.653 & 1.244 & 1.83\\
& max & 1.683 & 1.871 & 0.537 & 1.85 & 1.7 & 2.512 & 1.85 & 2.714\\
\hline
\multirow{3}{3mm}{\begin{sideways}\parbox{3mm}{80\%\ \ }\end{sideways}}
& average & 1.134 & 1.367 & 0.156 & 0.890 & 1.144 & 1.689 & 1.283 & 1.831\\
& min & 0.665 & 1.027 & 0.06 & 0.56 & 0.929 & 1.408 & 1.106 & 1.66\\
& max & 1.876 & 1.999 & 0.333 & 1.216 & 1.4 & 2.033 & 1.5 & 2.176\\
\hline
\end{tabular}
\label{all_results}
\end{small}
\end{center}
\end{table*}

\subsection{Predictive Power}
The predictive power of our model is measured by the $k>0$ next buildings for which we have ``good enough'' prediction of future build trees in:
$$ P(BuildTree^{t+k} | T=t, O_{1:N}=o_{1:N}, \lambda = 1)$$
``Good enough'' is measured by a distance $d$ to the actual build tree of the opponent that we tolerate. We used a set distance: $d(bt_1, bt_2) = \mathrm{card}(bt_1 \Delta bt_2) = \mathrm{card}((bt_1\bigcup bt_2) \backslash (bt_1\bigcap bt_2))$. One less or more building in the prediction is at a distance of $1$ from the actual build tree, the same buildings except for one difference is at a distance of $2$ (that would be $1$ is we used tree edit distance with substitution). We call $d(best,real)=$``best'' the distance between the most probable build tree and the one that actually happened. We call $d(bt,real)*P(bt)$=``mean'' the marginalized distance between what was inferred balanced (variable $bt$) by the probability of inferences ($P(bt)$). Note that this distance is always over \textbf{all} the build tree (and not only the next inference). This distance was taken into account only after the fourth (4th) building so that the first buildings would not penalize the prediction metric (the first building can not be predicted 4 buildings in advance). 

We used $d=1,2,3$, with $d=1$ we have a very strong sense of what the opponent is doing or will be doing, with $d=3$, we may miss one key building or have switched a tech path. We can see in Table~\ref{all_results} that with $d=1$ and without noise, our model predict in average more than one building in advance what the opponent will build next if we use only its best prediction, and almost \textbf{four} buildings in advance if we marginalize over all the predictions. Of course, if we accept more error, the predictive power (number of buildings ahead that our model is capable to predict) increases, up to $6.12$ for $d=3$ without noise.

\subsection{Robustness to Noise}
The robustness of our algorithm is measured by the quality of the predictions of the build trees for $k=0$ (reconstruction) or $k>0$ (prediction) with missing observations in:
$$ P(BuildTree^{t+k} | T=t, O_{1:N}=partial(o_{1:N}), \lambda = 1)$$
The ``reconstructive'' power (infer what has not been seen) ensues from the learning of our parameters from real data: even in the set of build trees that are possible, with regard to the game rules, only a few will be probable at a given time and/or with some key structures. Abiding by probability theory gives us consistency with regard to concurrent build tree.
This ``reconstructive'' power of our model is shown in Table~\ref{all_results} with $d$ (distance to actual building tree) for increasing noise at fixed $k=0$. 

\begin{figure}[htp]
\includegraphics[width=1.0\columnwidth]{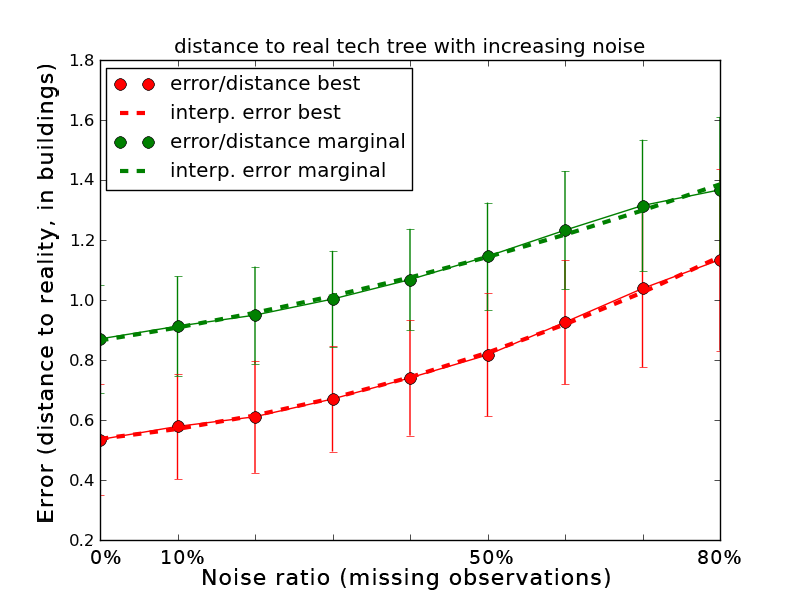}
\includegraphics[width=1.0\columnwidth]{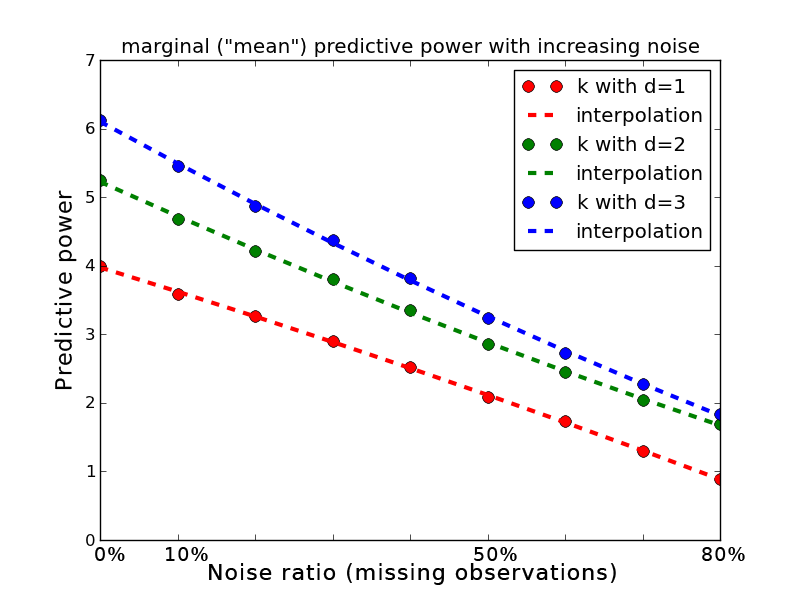}
\vspace{-0.5cm}
\caption{Evolution of our metrics with increasing noise, from 0 to 80\%. The top graphic shows the increase in distance between the predicted build tree, both most probable (``best'') and marginal (``mean'') and the actual one. The bottom graphic shows the decrease in predictive power: numbers of buildings ahead ($k$) for which our model predict a build tree closer than a fixed distance/error ($d$).}
\label{noise}
\end{figure}

Figure~\ref{noise} displays first (on top) the evolution of the error rate (distance to actual building) with increasing random noise (from 0\% to 80\%, no missing observations to 8 missing observations over 10). We consider that having an average distance to the actual build tree a little over $1$ for 80\% missing observations is a success. We think that this robustness is due to $P(T|BuildTree)$ being precise with the amount of data that we used. Secondly, Figure~\ref{noise} displays (at the bottom) the evolution of the predictive power (number of buildings ahead from the build tree that it can predict) with the same increase of noise.

\vspace{-0.3cm}

\section{Conclusions}
\subsection{Discussion and Perspectives}
Developing beforehand a RTS game AI that specifically deals with whatever strategies the players will come up is very hard. And even if game developers were willing to patch their AI afterwards, it would require a really modular design and a lot of work to treat each strategy. With our model, the AI can adapt to the evolutions in play by learning its parameters from the replay, it can dynamically adapt during the games by asking $P(BuilTree|Observations, Time, \lambda=1)$ and even $P(TechTree|Observations, Time, \lambda=1)$ if we add units and technology upgrades to buildings. This would allow for the bot to dynamically choose/change build orders and strategies.

This work can be extended by have a model for the two players (the bot/AI and the opponent):
$$P(BuildTree_{bot}, BuildTree_{op}, Obs_{op,1:N}, Time, \lambda)$$
So that we could ask this (new) model: $$P(BuildTree_{bot} | obs_{op, 1:N}, time, \lambda=1)$$
This would allow for simple and dynamic build tree adaptation to the opponent strategy (dynamic re-planning), by the inference path:
\begin{eqnarray*}
& P(BuildTree_{bot} | obs_{op, 1:N}, time, \lambda=1)\\
& \propto \sum_{BuildTree_{op}} P(BuildTree_{bot} | BuildTree_{op}) \ \mathrm{(learned)} \\
& \times P(BuildTree_{op}).P(o_{op, 1:N}) \ \mathrm{(priors)} \\
& \times P(\lambda | BuildTree_{op}, o_{op, 1:N}) \ \mathrm{(consistency)} \\
& \times P(time|BuildTree_{op}) \ \mathrm{(learned)}
\end{eqnarray*}
That way, one can ask ``what build/tech tree should I go for against what I see from my opponent'', which tacitly seeks the distribution on $BuildTree_{op}$ to break the complexity of the possible combinations of $Obs_{1:N}$. It is possible to \textit{not} marginalize over $BuildTree_{op}$, but consider only the most probable(s) $BuildTree_{op}$, for computing efficiency. A filter on $BuildTree_{bot}$ (as simple as $P(BuildTree_{bot}^t | BuildTree_{bot}^{t-1}$) can and should be added to prevent switching build orders or strategies too often.

The Bayesian model presented in this paper for opponent build tree prediction can be used in two main ways:
\begin{itemize}
\item as the corner stone of adaptive (to the opponent's dynamic strategies) RTS game AI:
\begin{itemize}
\item without noise in the case of built-in game AI (cheat).
\item with noise in the case of RTS AI tournaments (as AIIDE's) of matches against human players.
\end{itemize}
\item as a commentary assistant (null noise, prediction of tech trees), showing the probabilities of possible strategies as Poker commentary software do.
\end{itemize}

Finally, a hard problem is detecting the ``fake'' builds of very highly skilled players. Indeed, some pro-gamers have build orders which purposes are to fool the opponent into thinking that they are performing opening A while they are doing B. For instance they could ``take early gas'' leading the opponent to think they are going to do tech units, not gather gas and perform an early rush instead.

\subsection{Conclusion}

We presented a probabilistic model computing the distribution over build (or tech) trees of the opponent in a RTS game. The main contributions (with regard to \citeauthor{weber}) are the ability to deal with {\em partial} observations and {\em unsupervised} learning. This model yields high quality prediction results (up to $4$ buildings ahead with a total build tree distance less than $1$, see Table~\ref{all_results}) and shows a strong robustness to noise with a predictive power of $3$ buildings ahead with a build tree distance less than $1$ under 30\% random noise (a quality that we need for real setup/competitive games). It can be used in production thanks to its low computational (CPU) and memory footprint. Our implementation is free software and can be found online\footnote{\url{https://github.com/SnippyHolloW/OpeningTech/}}. We will use this model (or an upgraded version of it) in our StarCraft AI competition entry bot as it enables it to deal with the incomplete knowledge gathered from scouting.

\bibliographystyle{aaai}
\bibliography{aaai}

\begin{thebibliography}{}

\bibitem[\protect\citeauthoryear{Aha, Molineaux, and Ponsen}{2005}]{LTW}
Aha, D.~W.; Molineaux, M.; and Ponsen, M. J.~V.
\newblock 2005.
\newblock Learning to win: Case-based plan selection in a real-time strategy
  game.
\newblock In {\em ICCBR},  5--20.

\bibitem[\protect\citeauthoryear{Albrecht, Zukerman, and
  Nicholson}{1998}]{BayesianRecog}
Albrecht, D.~W.; Zukerman, I.; and Nicholson, A.~E.
\newblock 1998.
\newblock {B}ayesian models for keyhole plan recognition in an adventure game.
\newblock {\em User Modeling and User-Adapted Interaction} 8:5--47.

\bibitem[\protect\citeauthoryear{Beal}{2003}]{Beal}
Beal, M.~J.
\newblock 2003.
\newblock Variational algorithms for approximate {B}ayesian inference.
\newblock {\em PhD. Thesis}.

\bibitem[\protect\citeauthoryear{Bessi\`{e}re, Laugier, and
  Siegwart}{2008}]{PRDMSMS}
Bessi\`{e}re, P.; Laugier, C.; and Siegwart, R.
\newblock 2008.
\newblock {\em Probabilistic Reasoning and Decision Making in Sensory-Motor
  Systems}.
\newblock Springer Publishing Company, Incorporated.

\bibitem[\protect\citeauthoryear{Charniak and Goldman}{1993}]{BMPR}
Charniak, E., and Goldman, R.~P.
\newblock 1993.
\newblock A {B}ayesian model of plan recognition.
\newblock {\em Artificial Intelligence} 64(1):53--79.

\bibitem[\protect\citeauthoryear{Chung, Buro, and Schaeffer}{2005}]{Chung05}
Chung, M.; Buro, M.; and Schaeffer, J.
\newblock 2005.
\newblock {M}onte {C}arlo {P}lanning in {RTS} {G}ames.
\newblock In {\em CIG (IEEE)}.

\bibitem[\protect\citeauthoryear{Colas, Diard, and
  Bessi\`{e}re}{2010}]{Colas10}
Colas, F.; Diard, J.; and Bessi\`{e}re, P.
\newblock 2010.
\newblock Common {B}ayesian models for common cognitive issues.
\newblock {\em Acta Biotheoretica} 58:191--216.

\bibitem[\protect\citeauthoryear{Diard, Bessi\`{e}re, and
  Mazer}{2003}]{Diard03}
Diard, J.; Bessi\`{e}re, P.; and Mazer, E.
\newblock 2003.
\newblock A survey of probabilistic models using the {B}ayesian programming
  methodology as a unifying framework.
\newblock In {\em Conference on Computational Intelligence, Robotics and
  Autonomous Systems, CIRAS}.

\bibitem[\protect\citeauthoryear{Geib and Goldman}{2009}]{PHATT}
Geib, C.~W., and Goldman, R.~P.
\newblock 2009.
\newblock A probabilistic plan recognition algorithm based on plan tree
  grammars.
\newblock {\em Artificial Intelligence} 173:1101--1132.

\bibitem[\protect\citeauthoryear{Hoang, Lee-Urban, and
  Mu{\~n}oz-Avila}{2005}]{HTNPlanning}
Hoang, H.; Lee-Urban, S.; and Mu{\~n}oz-Avila, H.
\newblock 2005.
\newblock Hierarchical plan representations for encoding strategic game ai.
\newblock In {\em AIIDE},  63--68.

\bibitem[\protect\citeauthoryear{Hsieh and Sun}{2008}]{HsiehS08}
Hsieh, J.-L., and Sun, C.-T.
\newblock 2008.
\newblock Building a player strategy model by analyzing replays of real-time
  strategy games.
\newblock In {\em IJCNN},  3106--3111.

\bibitem[\protect\citeauthoryear{Jaynes}{2003}]{Jaynes}
Jaynes, E.~T.
\newblock 2003.
\newblock {\em Probability Theory: The Logic of Science}.
\newblock Cambridge University Press.

\bibitem[\protect\citeauthoryear{Kabanza \bgroup et al\mbox.\egroup
  }{2010}]{OBRecog}
Kabanza, F.; Bellefeuille, P.; Bisson, F.; Benaskeur, A.~R.; and Irandoust, H.
\newblock 2010.
\newblock Opponent behaviour recognition for real-time strategy games.
\newblock In {\em AAAI Workshops}.

\bibitem[\protect\citeauthoryear{Le~Hy \bgroup et al\mbox.\egroup
  }{2004}]{LeHy04}
Le~Hy, R.; Arrigoni, A.; Bessiere, P.; and Lebeltel, O.
\newblock 2004.
\newblock Teaching {B}ayesian behaviours to video game characters.
\newblock {\em Robotics and Autonomous Systems} 47:177--185.

\bibitem[\protect\citeauthoryear{Mishra, Onta{\~n}{\'o}n, and
  Ram}{2008}]{PlanRetrieval}
Mishra, K.; Onta{\~n}{\'o}n, S.; and Ram, A.
\newblock 2008.
\newblock Situation assessment for plan retrieval in real-time strategy games.
\newblock In {\em ECCBR},  355--369.

\bibitem[\protect\citeauthoryear{Ng and Jordan}{2001}]{Jordan}
Ng, A.~Y., and Jordan, M.~I.
\newblock 2001.
\newblock On discriminative vs. generative classifiers: A comparison of
  logistic regression and naive bayes.
\newblock In {\em NIPS},  841--848.

\bibitem[\protect\citeauthoryear{Onta\~{n}\'{o}n \bgroup et al\mbox.\egroup
  }{2007}]{OntanonCBR}
Onta\~{n}\'{o}n, S.; Mishra, K.; Sugandh, N.; and Ram, A.
\newblock 2007.
\newblock Case-based planning and execution for real-time strategy games.
\newblock In {\em Proceedings of the 7th International conference on Case-Based
  Reasoning: Case-Based Reasoning Research and Development}, ICCBR '07,
  164--178.
\newblock Springer-Verlag.

\bibitem[\protect\citeauthoryear{Onta{\~n}{\'o}n \bgroup et al\mbox.\egroup
  }{2008}]{CBR_Planning}
Onta{\~n}{\'o}n, S.; Mishra, K.; Sugandh, N.; and Ram, A.
\newblock 2008.
\newblock Learning from demonstration and case-based planning for real-time
  strategy games.
\newblock In Prasad, B., ed., {\em Soft Computing Applications in Industry},
  volume 226 of {\em Studies in Fuzziness and Soft Computing}. Springer Berlin
  / Heidelberg.
\newblock  293--310.

\bibitem[\protect\citeauthoryear{Ram\'{\i}rez and Geffner}{2009}]{Ramirez}
Ram\'{\i}rez, M., and Geffner, H.
\newblock 2009.
\newblock Plan recognition as planning.
\newblock In {\em Proceedings of the 21st international joint conference on
  Artifical intelligence},  1778--1783.
\newblock Morgan Kaufmann Publishers Inc.

\bibitem[\protect\citeauthoryear{Schadd, Bakkes, and
  Spronck}{2007}]{SchaddBS07}
Schadd, F.; Bakkes, S.; and Spronck, P.
\newblock 2007.
\newblock Opponent modeling in real-time strategy games.
\newblock In {\em GAMEON},  61--70.

\bibitem[\protect\citeauthoryear{Weber and Mateas}{2009}]{weber}
Weber, B.~G., and Mateas, M.
\newblock 2009.
\newblock A data mining approach to strategy prediction.
\newblock In {\em CIG (IEEE)}.

\end{thebibliography}

\end{document}